\documentclass[runningheads]{llncs}

\usepackage{eccv}

\makeatletter
\def\thanks#1{\protected@xdef\@thanks{\@thanks
        \protect\footnotetext{#1}}}
\makeatother
\usepackage{marvosym}

\usepackage{eccvabbrv}

\usepackage{graphicx}
\usepackage{booktabs}

\usepackage[accsupp]{axessibility}  

\usepackage{fontawesome}
\usepackage{multirow}
\usepackage{colortbl}
\usepackage{float}

\usepackage{wrapfig}

\def\etal{\textit{et al.}}
\def\eg{e.g.}
\def\ie{i.e.}
\definecolor{Gray}{gray}{0.9}

\newcommand{\hzw}[1]{\textcolor{black}{#1}}
\newcommand{\todo}[1]{\textcolor{black}{#1}}
\newcommand{\camera}[1]{\textcolor{black}{#1}}

\usepackage{hyperref}

\usepackage{orcidlink}

\begin{document}

\title{Unified Removal of Raindrops and Reflections: A New Benchmark and A Novel Pipeline} 

\titlerunning{Unified Removal of Raindrops and Reflections}


\author{Xingyu Liu\inst{1,2}\orcidlink{0009-0003-3815-4953} \and
Zewei He\inst{1,2}\textsuperscript{\Letter}\thanks{\textsuperscript{\Letter} Corresponding author (zeweihe@zju.edu.cn).}\orcidlink{0000-0003-4280-9708} \and
Yu Chen\inst{1,2}\orcidlink{0009-0002-6066-6990} \and
Chunyu Zhu\inst{3}\orcidlink{0000-0002-1290-6016} \and
Zixuan Chen\inst{4}\orcidlink{0009-0006-8283-3664} \and
Xing Luo\inst{2}\orcidlink{0000-0002-0879-1965} \and
Zhe-Ming Lu\inst{1,2}\orcidlink{0000-0003-1785-7847}}

\authorrunning{X.~Liu et al.}

\institute{Huanjiang Laboratory, Zhuji, China \and
School of Aeronautics and Astronautics, Zhejiang University, Hangzhou, China \and
Hangzhou Institute of Technology, Xidian University, Hangzhou, China \and
The Chinese University of Hong Kong, Hong Kong, China\\
\email{\{xingyu\_liu, zeweihe\}@zju.edu.cn}\\
\textbf{Project page}: \url{https://xingyuliu00.github.io/diffur3/}
}

\maketitle

\setcounter{footnote}{0} 

\begin{abstract}
When capturing images through glass surfaces or windshields on rainy days, raindrops and reflections frequently co-occur to significantly reduce the visibility of captured images. This practical problem lacks attention and needs to be resolved urgently. Prior de-raindrop, de-reflection, and all-in-one models have failed to address this composite degradation. To this end, we first formally define the unified removal of raindrops and reflections (UR$^3$) task for the first time and construct a real-shot dataset, namely RainDrop and ReFlection (RDRF), which provides a new benchmark with substantial, high-quality, diverse image pairs. Then, we propose \camera{an effective} diffusion-based framework (i.e., DiffUR$^3$) with several target designs to address this challenging task. By leveraging the powerful generative prior, DiffUR$^3$ successfully removes both types of degradations. Extensive experiments demonstrate that our method achieves state-of-the-art performance on our benchmark and on challenging in-the-wild images. 
\end{abstract}

\section{Introduction}
\label{sec:intro}
\hzw{On rainy days, raindrops and reflections frequently co-occur in autonomous driving scenarios, posing challenges for onboard visual recognition systems or vehicle cameras during recording.
Adherent raindrops and the reflections from camera side significantly reduce the visibility of captured images~\cite{You2016TPAMI}, and may lead to severe driving safety hazards.
Therefore, \textbf{U}nified \textbf{R}emoval of \textbf{R}aindrops and \textbf{R}eflections (UR$^3$) is a practical problem that requires urgent solution.
}

\begin{figure}[t]
	\centering
	\includegraphics[width=0.98\linewidth]{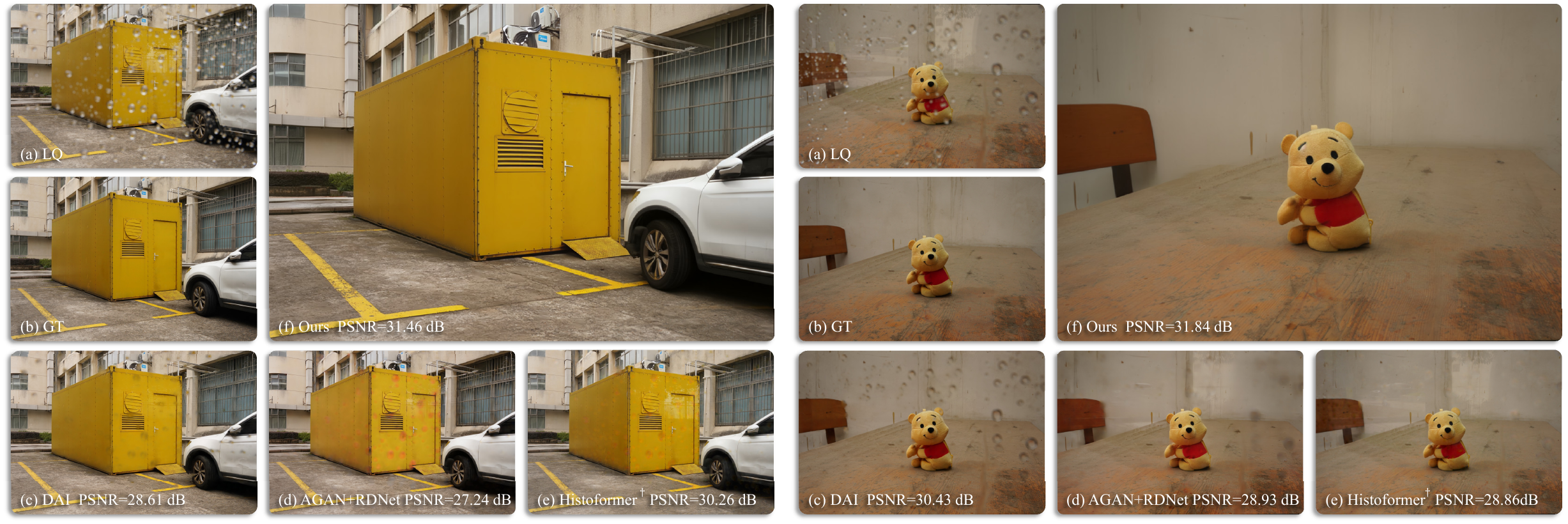}
	\caption{\hzw{We compare our DiffUR$^3$ pipeline with other methods on low-quality images with raindrops and reflections from our newly collected real-world benchmark. Specifically, (c) DAI \cite{Hu2026AAAI-DAI} is designed for reflection removal, (d) A cascaded method, and (e) A re-trained all-in-one method (\ie, Histoformer \cite{Sun2024ECCV-Hist} and $\dagger$ indicates re-trained on our dataset), (f) Our DiffUR$^3$ pipeline jointly removes both degradations in a single pass.}}
	\label{fig:fig1}
\end{figure}

\hzw{However, this task has not received significant attention and enough research.
We notice that the physical models of raindrop (i.e., $\textbf{I} = (1 - \textbf{A})\odot \textbf{B} + \textbf{A}\odot \textbf{R}_\textbf{d}$ from \cite{Quan2019ICCV}) and reflection (i.e., $\textbf{I} = (1 - \textbf{W})\odot \textbf{B} + \textbf{W}\odot \textbf{R}_\textbf{f}$ from \cite{Zheng2020CVPR-Rfmodel}) 
share a similar multiplicative compositing structure.
$\textbf{I}$ and $\textbf{B}$ denote the observed image and the background scene, respectively. $\textbf{R}_\textbf{d}$ and $\textbf{R}_\textbf{f}$ indicate the raindrop layer and the reflection layer, respectively. $\textbf{A}$ is the transparent matrix and $\textbf{W}$ is the reflective amplitude coefficient map.
When both degradations co-occur, the two processes are applied sequentially.
This yields a nested compositing model: $\textbf{I} = (1 - \textbf{A})\odot [(1 - \textbf{W})\odot \textbf{B} + \textbf{W}\odot \textbf{R}_\textbf{f}] + \textbf{A}\odot \textbf{R}_\textbf{d}$.
By expanding this expression, the cross-term $\textbf{A}\odot \textbf{W}\odot \textbf{R}_\textbf{f}$ couples both degradations.
This means the raindrop-related coefficient $\textbf{A}$ does not merely occlude the clean background $\textbf{B}$, but also modulates the reflection component $\textbf{W}\odot \textbf{R}_\textbf{f}$.
Therefore, the raindrop degradation and the reflection degradation are mutually coupled and interacted with each other. 
They do not simply superimpose upon each other.
The model designed for reflection removal can also partially eliminate raindrops (see in Fig.~\ref{fig:fig1}~(c)), and vice versa.}
Previously, researchers treated raindrop removal and reflection removal as two separate tasks ~\cite{Qian2018CVPR-AGAN,Quan2019ICCV,Shao2021TIP-UMAN,Hu2024NeurIPS-DSIT,Zhao2025CVPR-RDNet,Hu2026AAAI-DAI}.
Though these methods can achieve relatively good performance in removing the target type of degradation (\ie, raindrop or reflection) from a single image, they often fail to remove both types at the same time (see in Fig.~\ref{fig:fig1}~(c)).
\hzw{In addition, arranging raindrop removal and reflection removal in a cascaded manner essentially overlooks the correlation (see in Fig.~\ref{fig:fig1}~(d)).
In this work, we aim to finding a practical solution capable of simultaneously eliminating this raindrop-reflection composite degradation, thereby enhancing the clarity of captured images.}
We hope this endeavor can provide useful support for applications such as autonomous driving, photography, and video surveillance~\cite{Zhu2025IF,Zhu2025TGRS,Chen2024ECCV-PTTD,Chen2024TIP-DEANet}.

\hzw{
UR$^3$ is a fundamental but complex task. Its key challenges lie in the following three aspects.
(1) Lack of data: substantial image pairs are required for training.
Data synthesis represents a potential solution, yet the gap between synthetic and real-world data cannot be ignored.
Existing real-captured datasets \cite{Qian2018CVPR-AGAN,Zhu2024CVPR-RRW} normally contain only a single degradation.
Currently, there is no publicly available dataset with the raindrop-reflection composite degradation.
(2) Task interaction: treating raindrop removal and reflection removal as two isolated tasks often yields sub-optimal performance.
Recent all-in-one methods also demonstrate poor generalization performance (see in Fig.~\ref{fig:fig1}~(e)).
A new pipeline is required for this task.
(3) Void information: background scene is completely lost in regions with exceptionally large/dense raindrops or intense reflections.
These occluded regions are extremely challenging, somewhat analogous to the inpainting task.}

Trying to address these challenges, we first set up an image acquisition platform to collect corresponding data for addressing the UR$^3$ task. 
We collect a substantial number of image pairs to constitute our \textbf{R}ain\textbf{D}rop and \textbf{R}e\textbf{F}lection (RDRF) dataset.
Within each pair, one features a clean image and the other is a degraded image containing both raindrops and reflections. 
\hzw{To the best of our knowledge, RDRF is the first real-shot dataset with raindrop-reflection composite degradation.}
We hope RDRF dataset can contribute to the advancement and development of the UR$^3$ task, and benefit the entire community.

\camera{Then, we focus on proposing a qualified baseline for this new UR$^3$ task.}
By leveraging the powerful generative prior, we design \camera{an effective} multi-condition-controlled diffusion framework (\ie, DiffUR$^3$) to jointly remove both kinds of degradations.
\hzw{To the best of our knowledge, DiffUR$^3$ is the first pipeline for the UR$^3$ task, which contains several target designs.
(1) A Modulate\&Gate module is proposed to step-wisely align each condition with the noisy latent and adaptively select the effective components in the latent space. This simple yet effective module can enhance the control signals by modulating and mining the condition latents.
(2) We train an additional Fidelity Encoder to correct the distortions caused by the compression operation in the VAE encoder.
}


\section{Related work}
\subsection{Raindrop removal}
In the realm of raindrop removal, recent studies have explored diverse methodologies. Eigen \etal~\cite{Eigen2013ICCV} pioneered single-image raindrop removal using CNNs. Qian \etal~\cite{Qian2018CVPR-AGAN} introduced a generative adversarial network (GAN) to enhance raindrop removal. Transformer-based approaches like IDT \cite{Xiao2023TPAMI-IDT}, UDR-S$^2$Former \cite{Chen2023ICCV-UDR} and Histoformer \cite{Sun2024ECCV-Hist} have achieved superior performance. Meanwhile, the CCN \cite{Quan2021CVPR-CCN} adopts a unique approach by employing neural architecture search. More recently, diffusion-based methods like WeatherDiff \cite{Ozdenizci2023TPAMI} and T$^3$-DiffWeather \cite{Chen2024ECCV-T3Diffweather} have emerged, leveraging the generative capabilities of diffusion models to enhance raindrop removal.


\subsection{Reflection removal}
In the field of single image reflection removal, various advanced techniques have been proposed to address the ill-posed nature of separating superimposed transmission and reflection layers. Early methods such as CEILNet \cite{Fan2017ICCV-CEILNet} leverage edge information and deep learning. IBCLN \cite{Li2020CVPR-IBCLN} introduces a cascaded refinement strategy to iteratively enhance the estimations of transmission and reflection layers. More recent advancements, YTMT \cite{Hu2021NIPS-YTMT}, DSRNet \cite{Hu2023ICCV-DSRNet}, DSIT \cite{Hu2024NeurIPS-DSIT}, RDNet \cite{Zhao2025CVPR-RDNet}, and GFRRN \cite{Chen2026CVPR-GFRRN} employ dual-stream networks to enhance feature interaction. Further more, diffusion-based models like L-DiffER \cite{Hong2024ECCV-L-DiffER} and DAI \cite{Hu2026AAAI-DAI} also show their capabilities across a wide range of real-world scenarios.

\subsection{Datasets}
Existing real-world datasets for raindrop removal include AGAN \cite{Qian2018CVPR-AGAN}, RainDS \cite{Quan2021CVPR-CCN}, \todo{RobotCar-Rainy} \cite{Porav2019}, and Raindrop Clarity \cite{Jin2024ECCV-RaindropClarity}, these datasets provide low-quality images with raindrops and their corresponding ground truth images. 
Differently, Windshield \cite{Soboleva2021} contains degraded images along with their corresponding binary masks that indicate the raindrop-affected areas. For reflection removal task, it is noteworthy that synthetic data is commonly employed for training. Recently, some real-world datasets have been proposed, such as RRW \cite{Zhu2024CVPR-RRW} and DRR \cite{Hu2026AAAI-DAI}.
However, there are no existing datasets that specifically address the unified removal of raindrops and reflections, which is the focus of our work.


\begin{figure}[t]
	\centering
	\includegraphics[width=0.98\linewidth]{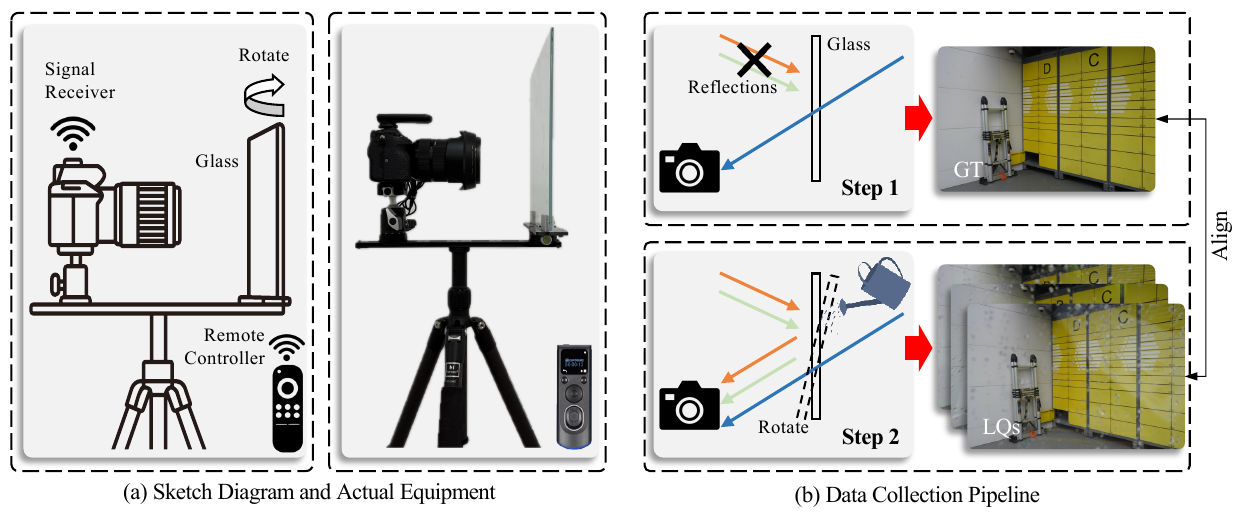}
	\caption{(a) Sketch diagram and actual equipment of our image acquisition platform. To suppress shutter-induced micro-vibrations which may potentially induce image misalignment, we implement a wireless triggering mechanism. It comprises a remote controller and a camera-mounted signal receiver, enabling contact-free shutter operation. (b) The data collection pipeline for our RDRF dataset. \faTimes~denotes light occlusion}
	\label{fig:fig2-collection}
\end{figure}

\begin{figure*}[t]
	\centering
	\includegraphics[width=0.98\linewidth]{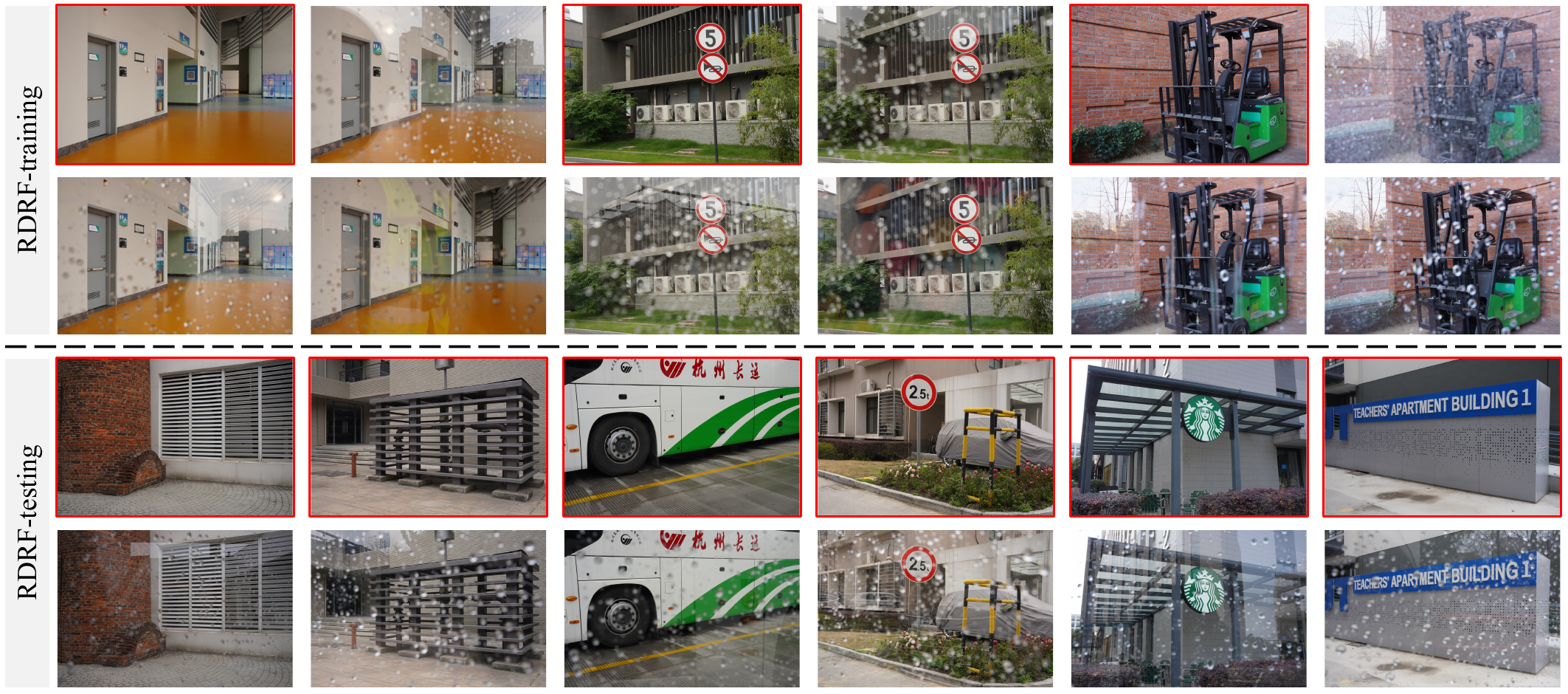}
	\caption{Our RDRF dataset comprises a diverse collection of scenes, each contains a ground truth and multiple low-quality images. As illustrated in this figure, the clean ground truths are highlighted in red boxes, while corresponding low-quality images are arranged around. We divide it into the training and testing subsets, ensuring no overlapping samples between them. Please zoom in on screen for a better view.}
	\label{fig:fig3}
\end{figure*}

\section{RDRF dataset}
Similar to most deep learning based methods, our task (\ie, unified removal of raindrops and reflections) requires a large number of degraded images with corresponding clean labels for training.
There are no existing training or testing datasets for this new task.
As shown in Fig.~\ref{fig:fig2-collection}~(a), we set up an image acquisition platform to collect our own \textbf{R}ain\textbf{D}rop and \textbf{R}e\textbf{F}lection (RDRF) dataset.
In our case, a substantial volume of image pairs are required, where each pair comprises two images with the identical background scene, yet one has a clean foreground and the other is corrupted by raindrops and reflections.

\subsection{Hardware}
Drawing inspirations from previous works of Zhu \etal~\cite{Zhu2024CVPR-RRW} and Li \etal~\cite{Li2024TPAMI}, our RDRF dataset is captured under real scenarios deliberately constructed in controlled environments.
For the hardware configuration, the camera is mounted on a tripod using an adjustable base, with the glass slab positioned in front of the lens.
We connect a signal receiver onto the camera, thereby enabling remote control of the shutter.
This wireless triggering mechanism can effectively avoid image misalignment caused by camera vibrations resulting from manual operation.
To ensure diversity, neither the camera nor the glass is fixed. 
They can be adjusted to simulate different shooting situations (\eg, camera-to-glass distances/angles).
In addition, we utilize two cameras (Sony ILCE-7RM4A and Nikon D7100) with zoom lens and choose different glass thicknesses (3 mm, 5 mm, and 8 mm) to further enhance diversity.

\subsection{Data collection pipeline}
Fig.~\ref{fig:fig2-collection}~(b) exhibits our data collection pipeline.
For step 1, we utilize a light-blocking box to suppress the reflections from the camera side (\faTimes~denotes light occlusion).
The obtained image is regarded as the ground-truth.
For step 2, we keep the background scenario and camera unchanged.
The light-blocking box is removed and the raindrops are created by spraying water onto the glass surface.
By randomly rotating the glass at different angles, we create varying reflections with different scenes and intensities.
For each scene, multiple images are captured as the low-quality ones.

Some samples are illustrated in Fig.~\ref{fig:fig3}.
As demonstrated in the dataset, our RDRF dataset comprises a comprehensive collection of scenes.
The raindrops are captured under diverse shapes and sizes (circular, elliptical, and irregular), ranging from sparse to dense.
Raindrop flow traces are also included.
In addition, the reflections are also captured with diverse reflection scenes, ranging from weak to strong.
All the images are captured in $4752\times 3168$ resolution to ensure high-quality.
In total, our RDRF dataset consists of 307 unique scenes. 
\textbf{The category distribution diagram can be found in the supplementary material.}
It is divided into a training set (216 scenes with 9003 image pairs) and a testing set (91 scenes with 277 image pairs).
\hzw{Note that, we capture 1 to 5 image pairs for each scene in the testing set.}

To further address the spatial misalignment caused by our hardware, we follow the procedures proposed in \cite{Wan2017ICCV}.
It starts by extracting SIFT \cite{Lowe2004IJCV-SIFT} key-points and descriptors, which are matched with L2 distance.
Using the matched key-points, a homography matrix is estimated via RANSAC \cite{Fischler1981-RANSAC} to handle outliers and find a robust geometric transformation.
The low-quality image is aligned to the ground truth by applying a perspective warp using the computed homography.


\hzw{
Previously, some valuable datasets are collected to address the raindrop removal task (\eg, Qian's dataset \cite{Qian2018CVPR-AGAN}, RainDS \cite{Quan2021CVPR-CCN}, Raindrop Clarity \cite{Jin2024ECCV-RaindropClarity}).
Their brilliant datasets has different focus with ours.
Some of their authors also noticed the reflection artifacts \cite{Qian2018CVPR-AGAN, Jin2024ECCV-RaindropClarity} and tried to avoid them during data collection stage.
Even so, there are still some reflection artifacts present in Qian's dataset \cite{Qian2018CVPR-AGAN}.
This also demonstrates that reflections and raindrops tend to occur simultaneously.
UR$^3$ is a practical task which required urgent solution.
Our RDRF dataset represents the first-of-its-kind contribution to the UR$^3$ task.
}

\section{Methodology}

Our RDRF dataset provides sufficient and diverse training data for UR$^3$ task.
Formally, given a low-quality (LQ) image $I_{lq} \in \mathbb{R}^{3\times H\times W}$ with both raindrops and reflections on it, a straightforward idea is to employ conventional restoration methods \cite{Chen2023CVPR-DRS,Sun2024ECCV-Hist} to directly learn the mapping function from the low-quality to the ground-truth.
However, their results are perceptually unsatisfying, because of the complexity of UR$^3$ task.
Instead, we try to utilize the powerful generative priors of the diffusion model as an effective solution.
In this way, UR$^3$ is regarded as a conditional image generation problem.

\begin{figure*}[t]
	\centering
	\includegraphics[width=0.98\linewidth]{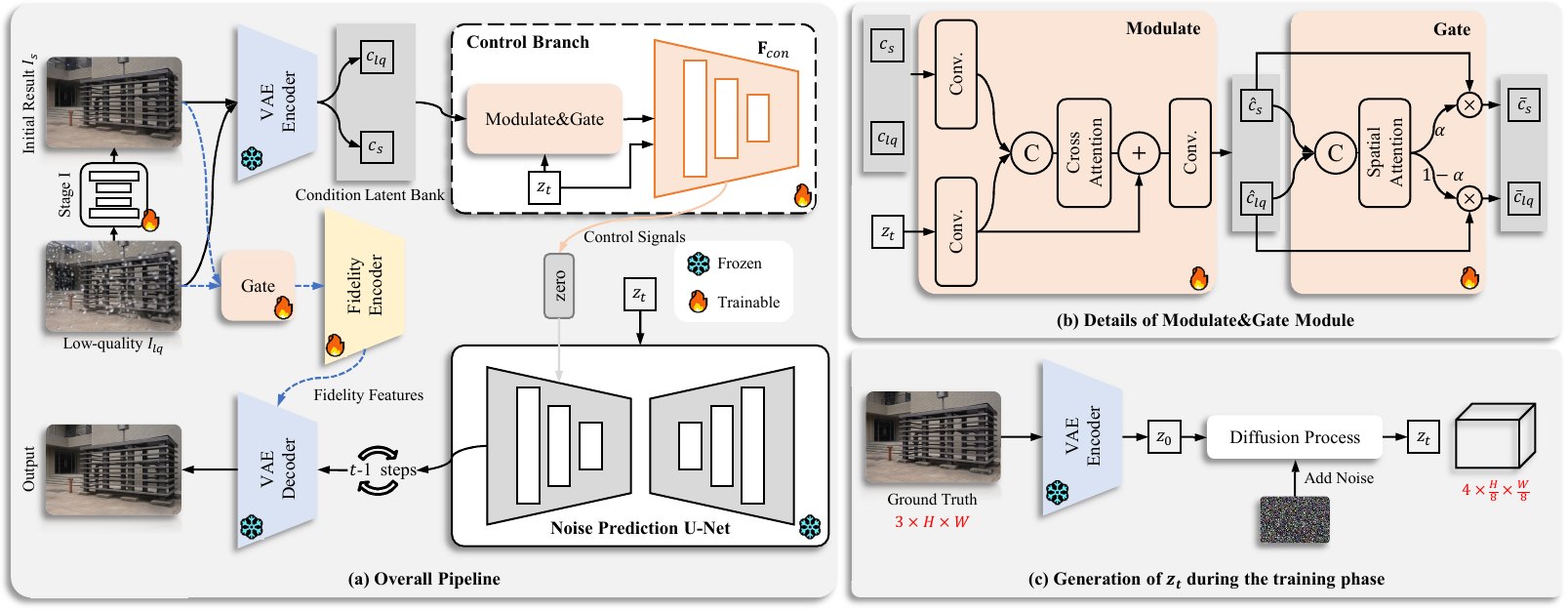}
	\caption{(a) Overall pipeline of our DiffUR$^3$ framework. Given a low-quality image $I_{lq}$, the restoration stage removes the undesired degradation to obtain the initial result $I_{s}$. Both $I_{lq}$ and $I_{s}$ are fed into the next stage as the condition images. We inject the \textbf{effective} condition information through a control branch, which outputs control signals for the noise prediction U-Net. (b) Details of the Modulate\&Gate module within the control branch. (c) The generation of noisy latent $z_{t}$ during the training phase. Note that the noisy latent starts from random Gaussian noise during the inference.}
	\label{fig:fig4}
\end{figure*}


Specifically, we design a two-stage network (\ie, DiffUR$^3$) shown in Fig.~\ref{fig:fig4}~(a), which consists of \textbf{(I) a restoration stage}, and \textbf{(II) a multi-condition generation stage}.
The restoration stage outputs an initial result $I_{s}$, which is then used as the condition image in the following generation stage.

Compared with DiffBIR framework, our DiffUR$^3$ has two major different designs: 
(1) Considering the fact that raindrops and reflections merely affect certain regions, some parts in the LQ image $I_{lq}$ can be regarded as clean. 
We also integrate $I_{lq}$ as one of the conditions during stage II.
A Modulate\&Gate module is designed to adaptively fuse the effective information from $I_{lq}$ and $I_{s}$.
(2) We design an additional Fidelity Encoder to correct the distortions caused by the compression in VAE. 
Details of our DiffUR$^3$ pipeline are described below.

\subsection{Restoration stage}
In the first stage, our aim is to remove some easy yet undesired degradations from LQ input $I_{lq}$.
The output image $I_{s}$ provides a reliable condition image for training the generation stage.
\begin{equation}
	I_{s} = \mathcal{RM}(I_{lq}),
\end{equation}
where $\mathcal{RM}(\cdot)$ denotes the restoration model.
In our implementation, we select the DRSformer \cite{Chen2023CVPR-DRS} as the restoration model in stage I due to its superior performance and generalization capability.

\subsection{Multi-condition generation stage}
Our multi-condition generation stage is based on the Stable Diffusion Model \cite{Rombach2022CVPR-LDM}, because its powerful generative prior can facilitate the restoration of regions that are challenging to be recovered in stage I through a conditional generation approach.
To achieve better efficiency and stabilized training, the pretrained VAE \cite{kingma2013ArXiv-VAE} encoder $\mathcal{E}$ is employed to encode the condition images into the latent space.
Both diffusion and denoising processes are performed in this space instead of the pixel space.
The main denoising network is a pretrained U-Net.
The denoising output is then converted back to the pixel space using the pretrained VAE decoder $\mathcal{D}$.

As mentioned above, for UR$^3$ task we argue LQ image $I_{lq}$ contains some clean information within certain regions \footnote{Unlike blind super-resolution and blind image denoising in \cite{Lin2024ECCV-DiffBIR}, where the entire LQ image is degraded.}. Therefore, both $I_{lq}$ and $I_{s}$ are encoded by the VAE encoder $\mathcal{E}$:
\begin{equation}
	c_{lq}, c_{s} = \mathcal{E}(I_{lq}, I_{s}),
\end{equation}
where $c_{lq} \in \mathbb{R}^{4\times \frac{H}{8}\times \frac{W}{8}}$ and $c_{s} \in \mathbb{R}^{4\times \frac{H}{8}\times \frac{W}{8}}$ denote the obtained condition latent from $I_{lq}$ and $I_{s}$, respectively.
Besides, the noisy latent $z_t$ is also embedded, since it has been proven to enhance image quality \cite{Lin2024ECCV-DiffBIR}.
The generation of $z_t$ is shown in Fig.~\ref{fig:fig4}~(c).

Similar to previous work \cite{Lin2024ECCV-DiffBIR}, we also inject the condition information via a control branch.
We make a trainable copy of the pretrained U-Net encoder and middle block (\ie, $\textbf{F}_{con}$ in Fig.~\ref{fig:fig4}~(a)), which receives condition information and then outputs control signals.
A normal solution is to add or concatenate $c_{lq}$, $c_{s}$ and $z_t$ before sending to $\textbf{F}_{con}$ \cite{Chen2025CVPR-FaithDiff,Ozdenizci2023TPAMI}.
However, we observe that the noisy latent $z_t$ varies at different time steps, yet the condition latent (\ie, $c_{lq}$ or $c_{s}$) remains unchanged.
Instead of direct addition or concatenation, we propose a more reasonable solution to modulate $c_{lq}$ and $c_{s}$ through $z_t$.
Since there are more than one condition latent, a gate mechanism is introduced to adaptively assign different spatial weights to $c_{lq}$ and $c_{s}$.

To this end, before entering the $\textbf{F}_{con}$, we design a Modulate\&Gate module which consists of a Modulate block and a Gate block. 
Fig.~\ref{fig:fig4}~(b) shows the details of our Modulate\&Gate module.
We describe them as below.

\subsubsection{Modulate block} 
Take $c_{s}$ as an example, $c_{lq}$ can be similarly derived.
First, both $c_{s}$ and $z_t$ individually pass through a convolutional layer to extract their features $f_{c} \in \mathbb{R}^{C\times \frac{H}{8}\times \frac{W}{8}}$ and $f_{z} \in \mathbb{R}^{C\times \frac{H}{8}\times \frac{W}{8}}$.
$C$ denotes the channel number of the extracted feature.
In our implementation, we set $C=32$.
Then, their concatenation result is fed into two consecutive transformer layers \cite{Vaswani2017NeurIPS} to perform the cross attention operation, which can facilitate the information interaction between $f_{c}$ and $f_{z}$.
Our cross attention operation aligns the dimensions of the output $f_{cross}$ and $f_{z}$ at the end.
Finally, we add $f_{cross} \in \mathbb{R}^{C\times \frac{H}{8}\times \frac{W}{8}}$ with $f_{z}$, and employ another convolutional layer to reduce the channel number back to $4$.
The formulations are as follows:
\begin{align}
	f_{c}, f_{z} &= Conv(c_{s}, z_{t}), \nonumber \\
	f_{cross} &= CrAttn([f_{c}, f_{z}]), \label{Eqn3}\\
	\hat{c}_{s} &= Conv(f_{cross} + f_{z}), \nonumber 
\end{align}
where $Conv(\cdot)$ denotes the convolutional layer, $CrAttn(\cdot)$ denotes the cross attention operation, $[\cdot,\cdot]$ denotes the concatenation, $\hat{c}_{s}$ is the modulated condition latent, and $\hat{c}_{lq}$ can be derived by replacing $c_{s}$ with $c_{lq}$ in Eqn.~\ref{Eqn3}.

\subsubsection{Gate block}
After obtaining $\hat{c}_{s}$ and $\hat{c}_{lq}$, we need to selectively extract the components that are beneficial to our DiffUR$^3$.
We concatenate them together, and then send to a spatial attention to generate a spatial weight $\alpha \in \mathbb{R}^{\frac{H}{8}\times \frac{W}{8}}$.
The spatial attention operation consists of two convolutional layers, one activation layer, and one sigmoid layer.
The formulations are as follow:
\begin{align}
	\alpha &= SpAttn([\hat{c}_{s}, \hat{c}_{lq}]), \nonumber \\
	\bar{c}_{s} &= \alpha \cdot \hat{c}_{s}, \\
	\bar{c}_{lq} &= (1-\alpha) \cdot \hat{c}_{lq}, \nonumber
\end{align}
where $SpAttn(\cdot)$ denotes the spatial attention operation, $\bar{c}_{s}$ and $\bar{c}_{lq}$ are the output condition latent variables.
Note that our Modulate\&Gate module is simple yet effective.
More sophisticated designs can be considered for better performance, which is not the focus of this work.
We concatenate $\bar{c}_{s}$, $\bar{c}_{lq}$, and the noisy latent $z_{t}$ together, and send them to $\textbf{F}_{con}$ for generating the control signals, which are added to the denoising U-Net via zero convolutions \cite{Zhang2023ICCV-ControlNet}.
At each time step, the noise prediction U-Net estimates the noise component and performs denoising on the noisy latent $z_{t}$.
During the inference phase, the noisy latent starts from random Gaussian noise and iteratively passes through the pretrained U-Net to estimate the clean latent $\hat{z}_{0}$.


\subsubsection{Fidelity encoder}
\begin{figure}[h]
	\centering
	\begin{minipage}[b]{.6\linewidth}
		\includegraphics[width=0.99\linewidth]{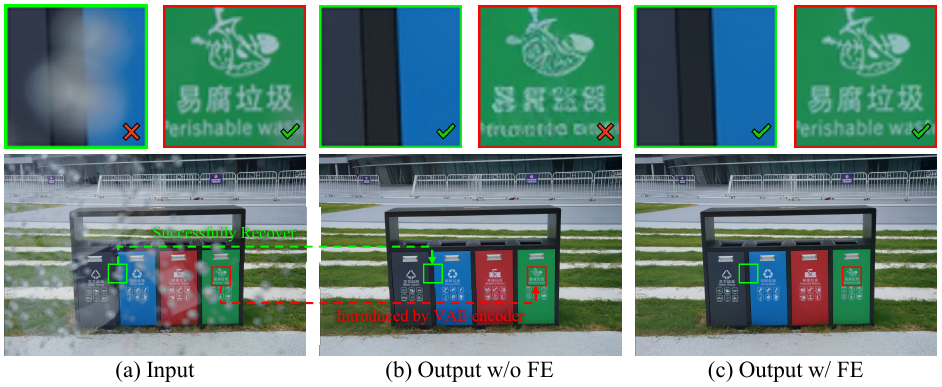}
		\caption{The motivation of employing an additional fidelity encoder.}
		\label{fig:fig5}
	\end{minipage}
	\quad
	\begin{minipage}[b]{.35\linewidth}
		\includegraphics[width=0.99\linewidth]{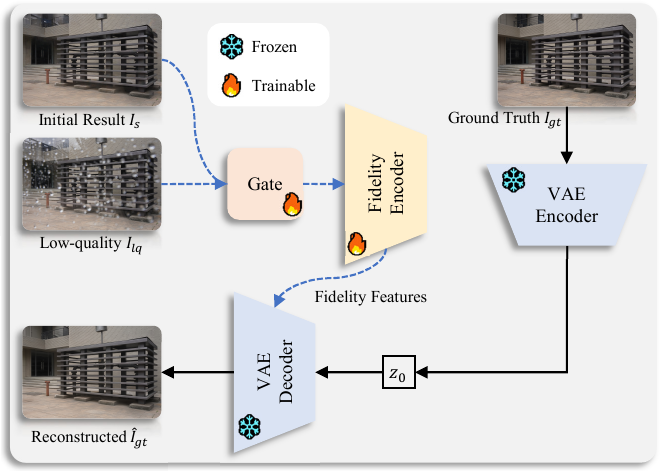}
		\caption{Training pipeline of our fidelity encoder.}
		\label{fig:fig6}
	\end{minipage}
\end{figure}

\hzw{
By decoding $\hat{z}_{0}$ back to the pixel space, we can obtain the final output image.
As shown in Fig.~\ref{fig:fig5}, we observe that while stable diffusion demonstrates superior performance in recovering the degradations caused by raindrops or reflections ({\color{green}green} box of Fig.~\ref{fig:fig5}~(b)), some unwanted distortions are introduced by the compression operation in the VAE encoder ({\color{red}red} box of Fig.~\ref{fig:fig5}~(b)).
The stable diffusion model lacks the ability to correct these kinds of distortions.
To deal with this issue and improve the fidelity of the generated results, we train an additional fidelity encoder (FE) inspired by \cite{Chang2023NeurIPS-L-CAD}.
The FE is proposed to extract multi-scale features from the initial result $I_s$ and the LQ image $I_{lq}$, which are not affected by the down-sampling, for preserving local structural semantics.
}


Fig.~\ref{fig:fig6} shows the training pipeline of our fidelity encoder, which shares the same architecture with VAE encoder \cite{kingma2013ArXiv-VAE} (besides the first convolutional layer).
To keep consistent with the control branch, both LQ image $I_{lq}$ and initial result $I_{s}$ are fed into the fidelity encoder through a Gate block to extract faithful features.
The extracted fidelity features are added to corresponding positions in VAE Decoder via zero convolutions.
Then, we encode the ground truth $I_{gt}$ via the pretrained VAE encoder to latent space, simulating the denoised latent \ie, $z_{0}$.
Finally, guided by the fidelity features, pretrained VAE decoder \cite{kingma2013ArXiv-VAE} converts the compressed latent $z_{0}$ to a reconstructed image $\hat{I}_{gt}$.
The whole pipeline is trained by minimizing a mean absolute error (\ie, $L_1$ loss) between $\hat{I}_{gt}$ and $I_{gt}$.

\subsubsection{Color correction}
\hzw{Diffusion models can occasionally exhibit color shifts \cite{Choi2022CVPR-color}.
To tackle this issue, some simple techniques (\eg, color normalization, wavelet color correction) have been explored before \cite{Wang2024IJCV-StableSR}.
We experimentally choose color normalization to align the mean and variance of the output with color reference $I_{s}$, and it is conducted patch-wisely.
Note that, $I_{lq}$ is not suitable here since the composite degradation may cause false correction.
}

\section{Experimental results}
\subsection{Implementation details and metrics}
For the training of stage I, we adopt DRSformer \cite{Chen2023CVPR-DRS} as the restoration model.
Compared with the original, we make certain simplifications to accelerate the computation in Stage I.
\{$N_0$,$N_1$,$N_2$,$N_3$,$N_4$\} are set to \{0,2,4,4,8\}, and the number of attention heads for sparse transformer blocks (\ie, STBs) in level 1 to level 4 are set to \{1,2,4,8\}. 
The initial channel $C$ is set to 16.
Patches of size $256\times 256$ are randomly cropped from the RDRF-training dataset, and horizontal and vertical flips are applied as the data augmentation techniques.
This model is trained with an initial learning rate of $3e^{-4}$ for the first 100K iterations, which will gradually reduced to $1e^{-6}$ using cosine annealing schedule \cite{He2019CVPR-Bag} during the remaining 200K iterations.

For stage II, the images in our RDRF-training dataset are randomly cropped into $640\times 640$ patches.
Horizontal and vertical flips, resizing, rotation are applied as the data augmentation techniques.
The control branch is trained with a batch size of 40 and with a fixed learning rate of $1e^{-4}$ for entire 50K iterations.
To accelerate the sampling process, we adopt a spaced DDPM sampling schedule \cite{Nichol2021ICML} which requires 50 sampling steps.
For the training of the additional fidelity encoder, we follow the training settings of stage II, except the batch size and number of iterations.

Note that the images in our RDRF-training dataset are firstly resized to a fixed resolution of $1080\times 720$, and we use AdamW \cite{AdamW} optimizer with default settings for all the training procedures.
We train the restoration model in stage I for 300k iterations (batch size = 4) on a single A6000 GPU.
Then we adopt the Stable Diffusion 2.1-base \cite{Rombach2022CVPR-LDM} as the generative prior, and train the control branch in stage II for 50k iterations (batch size = 40) on two A6000 GPUs.
The fidelity encoder is trained for 300k iterations (batch size = 6) on two A6000 GPUs.
We adopt three traditional metrics (PSNR, SSIM, LPIPS \cite{LPIPS}) and three no-reference image quality assessment metrics (MUSIQ \cite{MUSIQ}, CLIPIQA+ \cite{CLIPIQA}, HyperIQA \cite{HyberIQA}) to comprehensively evaluate our performance.

\subsection{Ablation study and discussions}
\subsubsection{Modulate\&Gate module}
First, we perform ablation study to validate the effectiveness of Modulate\&Gate module.
We employ the naive diffusion-based method which solely adopts $I_{lq}$ as the condition image (similar to \cite{Ozdenizci2023TPAMI}) and denote it as \textbf{Baseline 1}.
In addition, \textbf{Baseline 2} means only the $I_{s}$ is regarded as the condition image (similar to \cite{Lin2024ECCV-DiffBIR}).
Table~\ref{tab:tab1-ablation} summarizes the quantitative results.
\textbf{w/o Modulate\&Gate} Module means both $I_{lq}$ and $I_{s}$ are embedded as the condition images and fused by channel-wise concatenation in latent space.
We observe that our Modulate\&Gate module is critically important for our DiffUR$^3$, as omitting this component leads to an obvious performance drop.
Note that, both Modulate and Gate sub-modules are helpful for improving the performance.




\begin{table}[h]
	\centering
	\begin{minipage}[b]{.45\linewidth}
		\caption{Ablation study on Modulate\&Gate module.}
		\resizebox{\linewidth}{!}{
			\begin{tabular}{c|c|ccc}
				\toprule[1.2pt]
				Model & Condition	& PSNR$\uparrow$ & SSIM$\uparrow$ & LPIPS$\downarrow$\\
				\midrule \midrule
				Baseline 1  & $I_{lq}$	&  28.33 & 0.9271 & 0.0944 \\
				Baseline 2 & $I_{s}$	&  28.54 & 0.9273 & 0.0943  \\
				\midrule
				w/o Modulate\&Gate& $I_{lq},I_{s}$	& 28.60 & 0.9294  & 0.0899 \\
				w/o Modulate&$I_{lq},I_{s}$	&  29.11 & 0.9355 & 0.0832  \\
				w/o Gate&$I_{lq},I_{s}$	&  29.12 & 0.9344 & 0.0841  \\
				\midrule
				\rowcolor{Gray}DiffUR$^3$& $I_{lq},I_{s}$	& 29.41  & 0.9372  & 0.0813 \\
				\bottomrule[1.2pt]
			\end{tabular}
		}
		\label{tab:tab1-ablation}
	\end{minipage}
	\quad
	\begin{minipage}[b]{.35\linewidth}
		\caption{Ablation study on Fidelity Encoder.}
		\resizebox{\linewidth}{!}{
			\begin{tabular}{c|ccc}
				\toprule[1.2pt]
				Model	& PSNR$\uparrow$ & SSIM$\uparrow$ & LPIPS$\downarrow$\\
				\midrule \midrule
				Baseline	& 26.95  & 0.8042  & 0.1082 \\
				\midrule
				+CFW($w=0.5$) 	& 28.78 & 0.9120 & 0.0909 \\
				+CFW($w=1.0$) 	& 29.18 & 0.9320 & 0.0832  \\
				\midrule
				\rowcolor{Gray}+FE (DiffUR$^3$)	& 29.41 & 0.9372  & 0.0813  \\
				\bottomrule[1.2pt]
			\end{tabular}
		}
		\label{tab:tab2-ablation}
	\end{minipage}
\end{table}

\begin{figure}[t]
	\centering
	\includegraphics[width=0.9\linewidth]{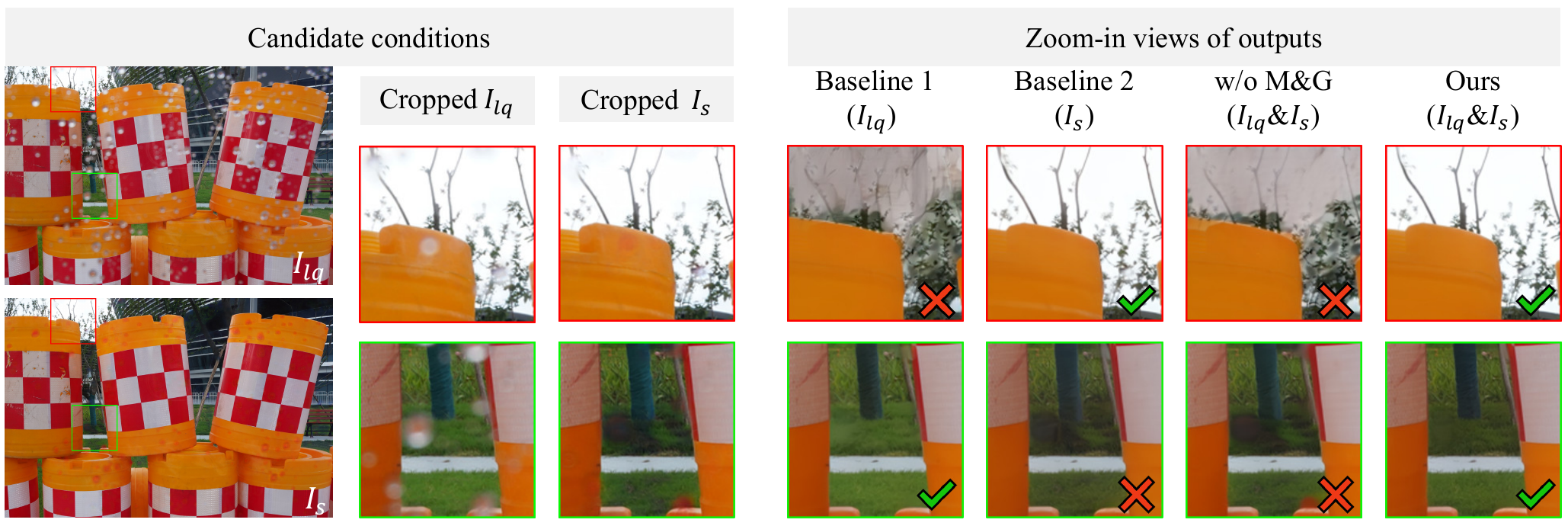}
	\caption{In-depth analysis on the function of our Modulate\&Gate (M\&G) module.}
	\label{fig:fig7}
\end{figure}

Further, an in-depth analysis is provided in Fig.~\ref{fig:fig7}.
\textbf{Baseline 1} outputs occasionally exhibit generation errors due to the inherent characteristics of diffusion model.
In contrast, $I_{s}$ is a better condition to avoid errors.
However, some restoration artifacts in $I_{s}$ may influence the \textbf{Baseline 2} outputs.
They demonstrate distinct advantages across different regions.
By considering $I_{lq}$ and $I_{s}$ together, simple channel-wise concatenation fails to systematically integrate their complementary strengths.
The introduced Modulate\&Gate module enables adaptive integration of information from dual condition images, thereby enhancing the model performance.


\subsubsection{Fidelity Encoder}
\begin{figure}[h]
	\centering
	\includegraphics[width=0.98\linewidth]{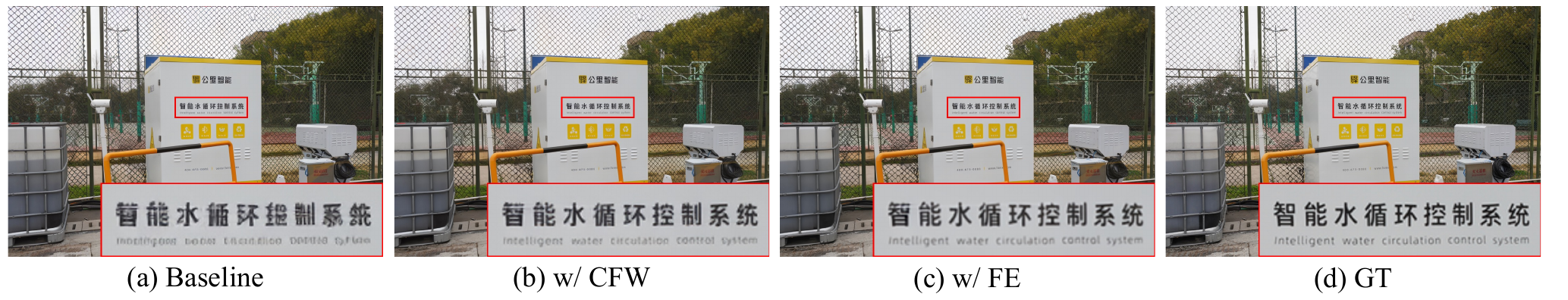}
	\caption{Visual results with CFW and FE.}
	\label{fig:fig-ab2}
\end{figure}

\hzw{Then, we perform ablation study to validate the effectiveness of the Fidelity Encoder (FE).
\textbf{Baseline} means directly decoding $\hat{z}_0$ to pixel space without fidelity features.
We observe a significant decline in fidelity, since the compression operation within the VAE encoder introduces considerable distortions (Fig.~\ref{fig:fig5}~(b)).
In addition, the Controllable Feature Wrapping (CFW) \cite{Wang2024IJCV-StableSR,Zhou2022NeurIPS-CodeFormer}, which leverages the encoder features to modulate corresponding decoder features for fidelity improvement, is also employed for comparison.}
\hzw{In CFW, there is a adjustable hyper-parameter $w$ to control the fidelity.
We choose $w=0.5$ and $w=1.0$.
The experimental results are listed in Table~\ref{tab:tab2-ablation}, and both CFW and FE can boost the fidelity.
Our FE ranks first in terms of PSNR, SSIM, and LPIPS.}
Some visual results are shown in Fig.~\ref{fig:fig-ab2}.
The texts recovered by our FE exhibit less distortions.
Note that, the training of our fidelity encoder is independent of both restoration and control branch, enabling its flexible application to various stable diffusion architectures.


\subsubsection{Color correction}
\begin{figure}[!h]
	\centering
	\includegraphics[width=0.98\linewidth]{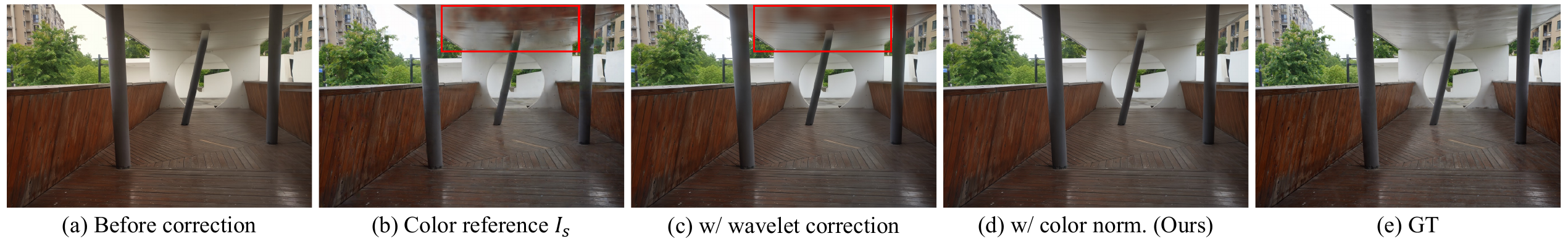}
	\caption{Color correction results with different methods.}
	\label{fig:fig9}
\end{figure}

\hzw{We experimentally test the color normalization and wavelet color correction in Fig.~\ref{fig:fig9}.
It can be observed that the wavelet method mistakenly introduces some artifacts from color reference $I_{s}$ in the roof region.
The color normalization corrects the color shift to make the result closer to the ground truth.
Our experimental results are inconsistent with \cite{Wang2024IJCV-StableSR}.
Since wavelet-based method introduces the low-frequency part from $I_{s}$, where may contain low-frequency artifacts caused by raindrops or reflections.
Therefore, we adopt the color normalization instead of the wavelet color correction in our case.
}

\subsection{Comparisons with state-of-the-art methods}

Since this work is the first exploration for unified removal of raindrops and reflections (UR$^3$) task.
There are no prior methods.
We employ two classical raindrop removal methods (\ie, AGAN \cite{Qian2018CVPR-AGAN}, UMAN \cite{Shao2021TIP-UMAN}), 
three classical reflection removal methods (\ie, RDNet \cite{Zhao2025CVPR-RDNet}, DSIT \cite{Hu2024NeurIPS-DSIT}, DAI \cite{Hu2026AAAI-DAI}), 
four cascaded methods,
and four all-in-one methods (\ie, Histoformer \cite{Sun2024ECCV-Hist}, WeatherDiff$_{64}$ \cite{Ozdenizci2023TPAMI}, DiffUIR \cite{Zheng2024CVPR-DiffUIR}, UniRestore \cite{Chen2025CVPR-UniRestore})
as the competitors.
We adopt their public models for calculating metrics.
In addition, we re-train the all-in-one methods by using their published codes.

\begin{table*}[t]
	\centering
	\caption{Benchmark results on our RDRF-testing dataset. We report PSNR, SSIM, LPIPS and three no-reference image quality assessment metrics (i.e., MUSIQ, CLIPIQA+, HyperIQA) to perform comprehensive comparisons. The \textbf{bold} and \underline{underline} indicate the best and second best. Superscript $\dagger$ means re-trained on RDRF.}
	\resizebox{0.99\textwidth}{!}{
		\begin{tabular}{c|c|c|cccccc}
			\toprule[1.2pt]
			\multirow{2}{*}{Type} & \multirow{2}{*}{Method} & \multirow{2}{*}{Venue} & \multicolumn{6}{c}{RDRF-testing} \\
			& & & PSNR$\uparrow$ & SSIM$\uparrow$ & LPIPS$\downarrow$ & MUSIQ$\uparrow$ & CLIPIQA+$\uparrow$ & HyperIQA$\uparrow$\\
			\midrule \midrule
			\multirow{2}{*}{\shortstack{Raindrop\strut\\removal}} 
			& AGAN \cite{Qian2018CVPR-AGAN} & CVPR2018 & 25.18 & 0.9209 & 0.1136 & 73.39 & 0.6241 & 0.6650 \\
			& UMAN \cite{Shao2021TIP-UMAN} & TIP2021 & 25.61 & 0.9222 & 0.1111 & 73.23 & 0.6262 & 0.6525 \\
			\midrule
			\multirow{3}{*}{\shortstack{Reflection\strut\\removal}} 
			& RDNet \cite{Zhao2025CVPR-RDNet} & CVPR2025 & 26.79 & 0.9169 & 0.1269 & 70.68 & 0.5983 & 0.6260 \\
			& DSIT \cite{Hu2024NeurIPS-DSIT} & NeurIPS2024 & 26.41 & 0.9128 & 0.1339 & 71.29 & 0.6058 & 0.6248 \\
			& DAI \cite{Hu2026AAAI-DAI} & AAAI2026 & 27.48 & 0.9251 & 0.1058 & 73.25 & 0.6335 & 0.6518 \\
			\midrule
			\multirow{6}{*}{Cascaded} 
			& AGAN+RDNet & - & 26.68 & 0.9291 & 0.0973 & 73.87 & 0.6550 & 0.6787 \\
			& RDNet+AGAN & - & 26.64 & 0.9268 & 0.0991 & 73.87 & 0.6453 & 0.6737 \\
			& AGAN+DSIT & - & 26.68 & 0.9283 & 0.0984 & \underline{74.05} & \underline{0.6661} & 0.6760 \\
			& DSIT+AGAN & - & 26.49 & 0.9243 & 0.1046 & 73.89 & 0.6536 & 0.6725 \\
			& UMAN+DAI & - & 27.50 & 0.9288 & 0.0973 & 74.01 & 0.6655 & \underline{0.6789} \\
			&DAI+UMAN & - & 27.44 & 0.9284 & 0.0966 & 73.95 & 0.6653 & 0.6766 \\
			\midrule
			\multirow{4}{*}{All-in-One} 
			& Histoformer \cite{Sun2024ECCV-Hist} & ECCV2024 & 25.74 & 0.9217 & 0.1214 & 72.34 & 0.5955 & 0.6450 \\
			& WeatherDiff$_{64}$ \cite{Ozdenizci2023TPAMI} & TPAMI2024 & 24.45 & 0.8936 & 0.1313 & 72.39 & 0.6383 & 0.6412 \\
			& DiffUIR \cite{Zheng2024CVPR-DiffUIR} & CVPR2024 & 23.32 & 0.8777 & 0.2190 & 68.39 & 0.5268 & 0.5901 \\
			& UniRestore \cite{Chen2025CVPR-UniRestore} & CVPR2025 & 22.77 & 0.8291 & 0.2952 & 66.71 & 0.5459 & 0.5123 \\
			\midrule
			\multirow{4}{*}{\shortstack{Re-trained\strut\\All-in-One}} 
			& Histoformer$^\dagger$ & - & 28.31 & 0.9315 & 0.0955 & 72.19 & 0.6221 & 0.6507 \\
			& WeatherDiff$_{64}$$^\dagger$ & - & 26.11 & 0.9130 & 0.1140 & 72.33 & 0.6471 & 0.6225 \\
			& DiffUIR$^\dagger$ & - & 28.39 & 0.9300 & 0.0991 & 71.83 & 0.6216 & 0.6311 \\
			& UniRestore$^\dagger$ & - & 25.66 & 0.8858 & 0.1631 & 73.01 & 0.6350 & 0.5994 \\
			\midrule
			& Stage I & - & \underline{28.82} & \underline{0.9354} & \underline{0.0925} & 73.48 & 0.6342 & 0.6648 \\
			\midrule
			\rowcolor{Gray}Ours& DiffUR$^3$ & - & \textbf{29.41} & \textbf{0.9372} & \textbf{0.0813} & \textbf{74.72} & \textbf{0.6705} & \todo{\textbf{0.7046}} \\
			\bottomrule[1.2pt]
		\end{tabular}
	}
	\label{tab:tab3}
\end{table*}

\begin{figure}[t]
	\centering
	\includegraphics[width=0.6\linewidth]{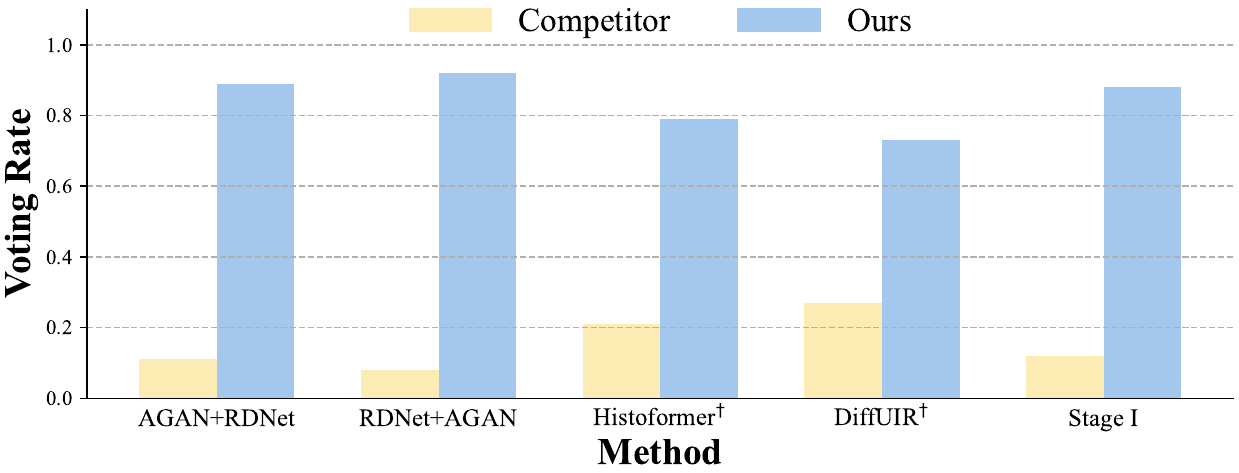}
	\caption{User study on RDRF-testing.}
	\label{fig:fig10-user-study}
\end{figure}

\begin{figure*}[h]
	\centering
	\includegraphics[width=0.99\linewidth]{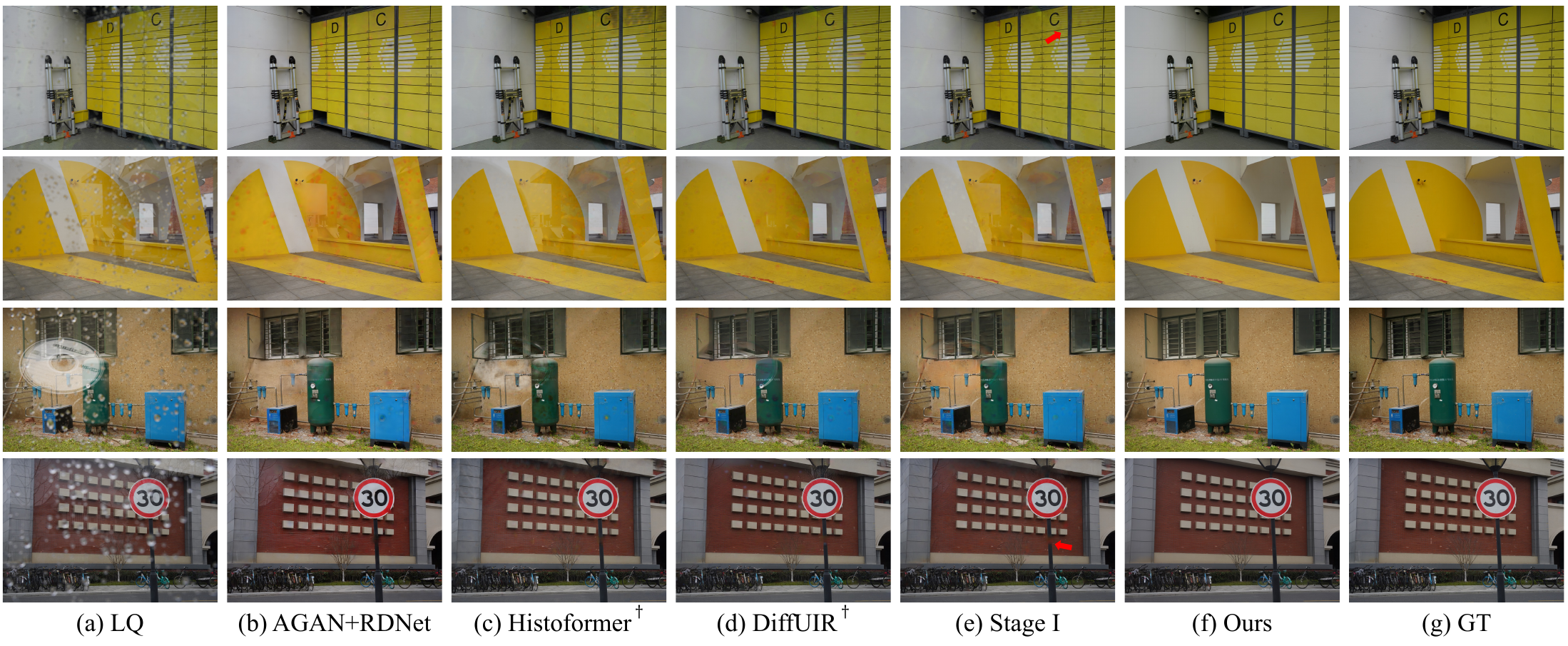}
	\caption{Visual results of various methods on our RDRF-testing. Superscript $\dagger$ means this method is re-trained on our RDRF-traing dataset. Please check and zoom in on screen for a better view.}
	\label{fig:fig11}
\end{figure*}

\begin{figure*}[h]
	\centering
	\includegraphics[width=0.99\linewidth]{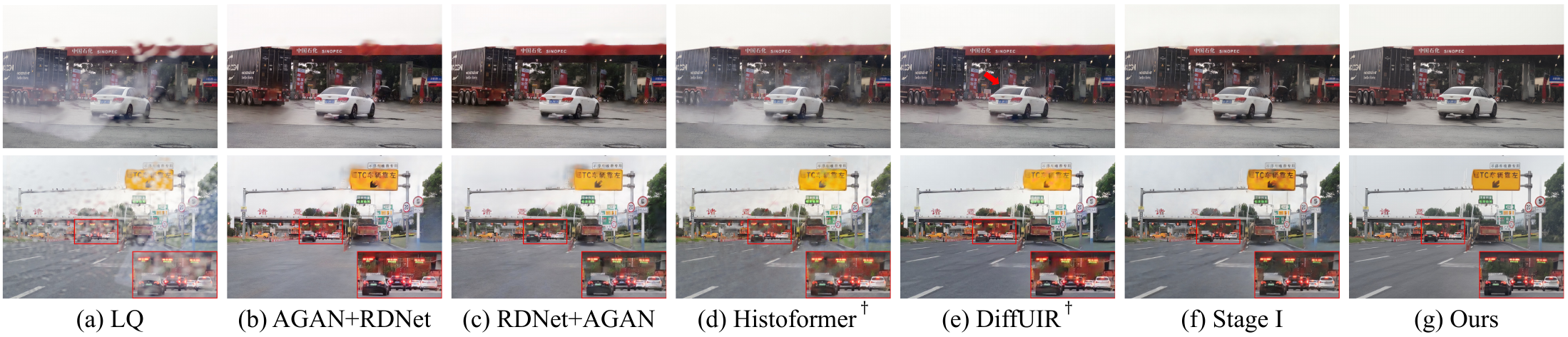}
	\caption{Visual results of various methods on our RDRF-wild dataset. Please check and zoom in on screen for a better view.}
	\label{fig:fig12}
\end{figure*}

We conduct a user study to evaluate our DiffUR$^3$ subjectively against five methods. 
Specifically, we randomly select 50 images from our RDRF-testing dataset and invite 20 experts with image restoration background as volunteers.
For every image, each expert is asked to compare the result of our DiffUR$^3$ with the alternatives one by one.
For each comparison, the observers are demanded to choose the favored one after at least 10 seconds of observation.
Afterward, we statistic the percentage of certain method to be selected.
The statistical results in Fig.~\ref{fig:fig10-user-study} indicates that our DiffUR$^3$ is more favored by the invited experts.

\hzw{Table~\ref{tab:tab3} shows the quantitative results on our RDRF-testing dataset.
Note that, our DiffUR$^3$ ranks top among all metrics.
Though the output from stage I exhibits relatively higher PSNR and SSIM values, it still suffers from certain distortions and residual degradations (Fig.~\ref{fig:fig11}~(e)), resulting in unsatisfactory visual quality and lower no-reference scores. 
In contrast, stage II and Fidelity Encoder effectively refines the output and produces significantly higher-quality results.}


In addition, some visual comparisons of our DiffUR$^3$ and the competitors are provided in Fig.~\ref{fig:fig11}.
It is worth mentioning that the results of our DiffUR$^3$ are closer to the ground truth with less degradation residuals and artifacts than the alternatives.
We further evaluate the generalization capability of our DiffUR$^3$ on a raindrop-only dataset~\cite{Qian2018CVPR-AGAN} (single degradation situation).
\textbf{More details are provided in the supplementary material.}
Since real-world driving scenarios in rainy weather are very challenging, we capture some testing images to form a RDRF-wild dataset.
Fig.~\ref{fig:fig12} shows the visual comparisons on RDRF-wild.
For this situation, the whole scene is more complicated than the controlled one.
Our DiffUR$^3$ can still obtain appealing results.
This indicates the robustness and generalization capability of our approach in real-world applications.
\textbf{More visual results can be found in the supplementary material.}




\section{Conclusion}
This work introduces a pioneering approach to the challenging task of UR$^3$. 
By establishing the first dedicated RDRF dataset and proposing \camera{an effective} diffusion-based framework (DiffUR$^3$), we successfully address the limitations of previous methods that treat raindrop and reflection removal as separate tasks. 
Our two-stage pipeline, incorporating a restoration stage and a multi-condition generation stage, effectively leverages generative priors to remove both types of degradations simultaneously. 
Extensive experiments demonstrate the superiority of our approach over SOTA methods. 
The RDRF dataset and DiffUR$^3$ framework contribute significantly to the advancement of the UR$^3$ task, offering valuable resources for future research.

\section*{Acknowledgment}
This work was supported in part by National Natural Science Foundation of China under Grant No. 52305590, Zhejiang Provincial Natural Science Foundation of China under Grant No. LQ24F010004, and Tianshan Talent Cultivation Plan - Science and Technology Innovation Team Project under Grant No. 2024TSYCTD0011.

\bibliographystyle{splncs04}
\bibliography{main}

@String(CVPR= {IEEE Conf. Comput. Vis. Pattern Recog.})

@String(ICCV= {Int. Conf. Comput. Vis.})

@String(ECCV= {Eur. Conf. Comput. Vis.})

@String(ICLR = {Int. Conf. Learn. Represent.})

@String(AAAI = {AAAI})

@String(CVPR  = {CVPR})

@String(ICCV  = {ICCV})

@String(ECCV  = {ECCV})

@String(ICLR  = {ICLR})

@inproceedings{Qian2018CVPR-AGAN,
	author = {Qian, Rui and Tan, Robby T. and Yang, Wenhan and Su, Jiajun and Liu, Jiaying},
	booktitle = {CVPR},
	pages = {2482--2491},
	title = {{Attentive Generative Adversarial Network for Raindrop Removal from a Single Image}},
	year = {2018}
}

@inproceedings{Quan2019ICCV,
	author = {Quan, Yuhui and Deng, Shijie and Chen, Yixin and Ji, Hui},
	booktitle = {ICCV},
	pages = {2463--2471},
	title = {{Deep learning for seeing through window with raindrops}},
	year = {2019}
}

@inproceedings{Hu2026AAAI-DAI,
	author = {Hu, Jichen and Yang, Chen and Zhou, Zanwei and Fang, Jiemin and Yang, Xiaokang and Tian, Qi and Shen, Wei},
	booktitle = {AAAI},
	title = {{Dereflection Any Image with Diffusion Priors and Diversified Data}},
	pages = {4860--4868},
	year = {2026}
}

@inproceedings{Hu2024NeurIPS-DSIT,
	title={Single image reflection separation via dual-stream interactive transformers},
	author={Hu, Qiming and Wang, Hainuo and Guo, Xiaojie},
	booktitle={NeurIPS},
	volume={37},
	pages={55228--55248},
	year={2024}
}

@inproceedings{Zhao2025CVPR-RDNet,
	title={Reversible decoupling network for single image reflection removal},
	author={Zhao, Hao and Li, Mingjia and Hu, Qiming and Guo, Xiaojie},
	booktitle={CVPR},
	pages={26430--26439},
	year={2025}
}

@inproceedings{kingma2013ArXiv-VAE,
  author       = {Diederik P. Kingma and Max Welling},
  title        = {Auto-Encoding Variational Bayes},
  booktitle    = {ICLR},
  year         = {2014},
}

@inproceedings{Chen2023CVPR-DRS,
	author = {Chen, Xiang and Li, Hao and Li, Mingqiang and Pan, Jinshan},
	booktitle = {CVPR},
	pages = {5896--5905},
	title = {{Learning A Sparse Transformer Network for Effective Image Deraining}},
	year = {2023}
}

@inproceedings{Sun2024ECCV-Hist,
	author = {Sun, Shangquan and Ren, Wenqi and Gao, Xinwei and Wang, Rui and Cao, Xiaochun},
	booktitle = {ECCV},
	title = {{Restoring Images in Adverse Weather Conditions via Histogram Transformer}},
	pages = {111--129},
	year = {2024}
}

@inproceedings{Lin2024ECCV-DiffBIR,
	author = {Lin, Xinqi and He, Jingwen and Chen, Ziyan and Lyu, Zhaoyang and Dai, Bo and Yu, Fanghua and Qiao, Yu and Ouyang, Wanli and Dong, Chao},
	booktitle = {ECCV},
	pages = {430--448},
	title = {{DiffBIR: Toward Blind Image Restoration with Generative Diffusion Prior}},
	year = {2024}
}

@inproceedings{Rombach2022CVPR-LDM,
	title={High-resolution image synthesis with latent diffusion models},
	author={Rombach, Robin and Blattmann, Andreas and Lorenz, Dominik and Esser, Patrick and Ommer, Bj{\"o}rn},
	booktitle={CVPR},
	pages={10684--10695},
	year={2022}
}

@inproceedings{Chen2025CVPR-FaithDiff,
	author = {Chen, Junyang and Pan, Jinshan and Dong, Jiangxin},
	booktitle = {CVPR},
	pages = {28188--28197},
	title = {{FaithDiff: Unleashing Diffusion Priors for Faithful Image Super-resolution}},
	year = {2025}
}

@inproceedings{Vaswani2017NeurIPS,
	author = {Vaswani, Ashish and Shazeer, Noam and Parmar, Niki and Uszkoreit, Jakob and Jones, Llion and Gomez, Aidan N and Kaiser, \L ukasz and Polosukhin, Illia},
	booktitle = {NeurIPS},
	title = {Attention is All you Need},
	volume = {30},
	pages = {5998--6008},
	year = {2017}
}

@article{Ozdenizci2023TPAMI,
	author = {{\"{O}}zdenizci, Ozan and Legenstein, Robert},
	journal = {IEEE Transactions on Pattern Analysis and Machine Intelligence},
	number = {8},
	pages = {10346--10357},
	title = {{Restoring Vision in Adverse Weather Conditions With Patch-Based Denoising Diffusion Models}},
	volume = {45},
	year = {2023}
}

@inproceedings{Zhang2023ICCV-ControlNet,
	title={Adding conditional control to text-to-image diffusion models},
	author={Zhang, Lvmin and Rao, Anyi and Agrawala, Maneesh},
	booktitle={ICCV},
	pages={3836--3847},
	year={2023}
}

@inproceedings{Chang2023NeurIPS-L-CAD,
	author = {Chang, Zheng and Weng, Shuchen and Zhang, Peixuan and Li, Yu and Li, Si and Shi, Boxin},
	booktitle = {NeurIPS},
	pages = {1--13},
	title = {{L-CAD: Language-based Colorization with Any-level Descriptions using Diffusion Priors}},
	volume = {36},
	year = {2023}
}

@inproceedings{Wan2017ICCV,
	author = {Wan, Renjie and Shi, Boxin and Duan, Ling-Yu and Tan, Ah-Hwee and Kot, Alex C.},
	booktitle = {ICCV},
	pages = {3922--3930},
	title = {{Benchmarking Single-Image Reflection Removal Algorithms}},
	year = {2017}
}

@article{Lowe2004IJCV-SIFT,
	title={Distinctive image features from scale-invariant keypoints},
	author={Lowe, David G},
	journal={International journal of computer vision},
	volume={60},
	number={2},
	pages={91--110},
	year={2004}
}

@article{Fischler1981-RANSAC,
	title={Random sample consensus: a paradigm for model fitting with applications to image analysis and automated cartography},
	author={Fischler, Martin A and Bolles, Robert C},
	journal={Communications of the ACM},
	volume={24},
	number={6},
	pages={381--395},
	year={1981}
}

@inproceedings{Jin2024ECCV-RaindropClarity,
	author = {Jin, Yeying and Li, Xin and Wang, Jiadong and Zhang, Yan and Zhang, Malu},
	booktitle = {ECCV},
	title = {{Raindrop Clarity: A Dual-Focused Dataset for Day and Night Raindrop Removal}},
	pages = {1--17},
	year = {2024}
}

@article{You2016TPAMI,
	author = {You, Shaodi and Tan, Robby T. and Kawakami, Rei and Ikeuchi, Katsushi},
	journal = {IEEE Transactions on Pattern Analysis and Machine Intelligence},
	number = {9},
	pages = {1721--1733},
	title = {{Adherent Raindrop Detection and Removal in Video}},
	volume = {38},
	year = {2016}
}

@article{Shao2021TIP-UMAN,
	author = {Shao, Ming Wen and Li, Le and Meng, De Yu and Zuo, Wang Meng},
	journal = {IEEE Transactions on Image Processing},
	pages = {4828--4839},
	title = {{Uncertainty Guided Multi-Scale Attention Network for Raindrop Removal from a Single Image}},
	volume = {30},
	year = {2021}
}

@inproceedings{Chen2023ICCV-UDR,
	author = {Chen, Sixiang and Ye, Tian and Bai, Jinbin and Chen, Erkang and Shi, Jun and Zhu, Lei},
	booktitle = {ICCV},
	pages = {13060--13071},
	title = {{Sparse Sampling Transformer with Uncertainty-Driven Ranking for Unified Removal of Raindrops and Rain Streaks}},
	year = {2023}
}

@inproceedings{Eigen2013ICCV,
	author = {Eigen, David and Krishnan, Dilip and Fergus, Rob},
	booktitle = {ICCV},
	pages = {633--640},
	title = {{Restoring an image taken through a window covered with dirt or rain}},
	year = {2013}
}

@inproceedings{Quan2021CVPR-CCN,
	author = {Quan, Ruijie and Yu, Xin and Liang, Yuanzhi and Yang, Yi},
	booktitle = {CVPR},
	pages = {9143--9152},
	title = {{Removing Raindrops and Rain Streaks in One Go}},
	year = {2021}
}

@article{Xiao2023TPAMI-IDT,
	author = {Xiao, Jie and Fu, Xueyang and Liu, Aiping and Wu, Feng and Zha, Zheng Jun},
	journal = {IEEE Transactions on Pattern Analysis and Machine Intelligence},
	number = {11},
	pages = {12978--12995},
	title = {{Image De-Raining Transformer}},
	volume = {45},
	year = {2023}
}

@inproceedings{Chen2024ECCV-T3Diffweather,
	author = {Chen, Sixiang and Ye, Tian and Zhang, Kai and Xing, Zhaohu and Lin, Yunlong and Zhu, Lei},
	booktitle = {ECCV},
	pages = {95--115},
	title = {{Teaching Tailored to Talent: Adverse Weather Restoration via Prompt Pool and Depth-Anything Constraint}},
	year = {2024}
}

@inproceedings{Fan2017ICCV-CEILNet,
	author = {Fan, Qingnan and Yang, Jiaolong and Hua, Gang and Chen, Baoquan and Wipf, David},
	booktitle = {ICCV},
	pages = {3238--3247},
	title = {{A Generic Deep Architecture for Single Image Reflection Removal and Image Smoothing}},
	year = {2017}
}

@inproceedings{Li2020CVPR-IBCLN,
	author = {Li, Chao and Yang, Yixiao and He, Kun and Lin, Stephen and Hopcroft, John E.},
	booktitle = {CVPR},
	pages = {3562--3571},
	title = {{Single image reflection removal through cascaded refinement}},
	year = {2020}
}

@inproceedings{Hu2021NIPS-YTMT,
	author = {Hu, Qiming and Guo, Xiaojie},
	booktitle = {NeurIPS},
	pages = {24683--24694},
	title = {{Trash or Treasure? An Interactive Dual-Stream Strategy for Single Image Reflection Separation}},
	volume = {30},
	year = {2021}
}

@inproceedings{Hu2023ICCV-DSRNet,
	author = {Hu, Qiming and Guo, Xiaojie},
	booktitle = {ICCV},
	pages = {13092--13101},
	title = {{Single Image Reflection Separation via Component Synergy}},
	year = {2023}
}

@inproceedings{Hong2024ECCV-L-DiffER,
	author = {Hong, Yuchen and Zhong, Haofeng and Weng, Shuchen and Liang, Jinxiu and Shi, Boxin},
	booktitle = {ECCV},
	pages = {58--76},
	title = {{L-DiffER: Single Image Reflection Removal with Language-Based Diffusion Model}},
	year = {2024}
}

@inproceedings{Soboleva2021,
	author = {Soboleva, Vera and Shipitko, Oleg},
	booktitle = {IEEE Symposium Series on Computational Intelligence},
	pages = {1--7},
	title = {{Raindrops on Windshield: Dataset and Lightweight Gradient-Based Detection Algorithm}},
	year = {2021}
}

@inproceedings{Porav2019,
	author = {Porav, Horia and Bruls, Tom and Newman, Paul},
	booktitle = {ICRA},
	pages = {7087--7093},
	title = {{I can see clearly now: Image restoration via de-raining}},
	year = {2019}
}

@inproceedings{Zhu2024CVPR-RRW,
	author = {Zhu, Yurui and Fu, Xueyang and Jiang, Peng-Tao and Zhang, Hao and Sun, Qibin and Chen, Jinwei and Zha, Zheng-Jun and Li, Bo},
	booktitle = {CVPR},
	pages = {25468--25478},
	title = {{Revisiting Single Image Reflection Removal In the Wild}},
	year = {2024}
}

@article{Li2024TPAMI,
	author = {Li, Yizhou and Monno, Yusuke and Okutomi, Masatoshi},
	journal = {IEEE Transactions on Pattern Analysis and Machine Intelligence},
	number = {12},
	pages = {10748--10762},
	title = {{Dual-Pixel Raindrop Removal}},
	volume = {46},
	year = {2024}
}

@inproceedings{MUSIQ,
	title={Musiq: Multi-scale image quality transformer},
	author={Ke, Junjie and Wang, Qifei and Wang, Yilin and Milanfar, Peyman and Yang, Feng},
	booktitle={ICCV},
	pages={5148--5157},
	year={2021}
}

@inproceedings{HyberIQA,
	author = {Su, Shaolin and Yan, Qingsen and Zhu, Yu and Zhang, Cheng and Ge, Xin and Sun, Jinqiu and Zhang, Yanning},
	booktitle = {CVPR},
	pages = {3664--3673},
	title = {{Blindly Assess Image Quality in the Wild Guided by a Self-Adaptive Hyper Network}},
	year = {2020}
}

@inproceedings{LPIPS,
	title={The unreasonable effectiveness of deep features as a perceptual metric},
	author={Zhang, Richard and Isola, Phillip and Efros, Alexei A and Shechtman, Eli and Wang, Oliver},
	booktitle={CVPR},
	pages={586--595},
	year={2018}
}

@inproceedings{CLIPIQA,
	author = {Jianyi Wang and Kelvin C. K. Chan and Chen Change Loy},
	title = {Exploring {CLIP} for Assessing the Look and Feel of Images},
	booktitle = {AAAI},
	pages = {2555--2563},
	year = {2023}
}

@inproceedings{AdamW,
  author       = {Ilya Loshchilov and Frank Hutter},
  title        = {Decoupled Weight Decay Regularization},
  booktitle    = {ICLR},
  year         = {2019}
}

@inproceedings{He2019CVPR-Bag,
	title={Bag of tricks for image classification with convolutional neural networks},
	author={He, Tong and Zhang, Zhi and Zhang, Hang and Zhang, Zhongyue and Xie, Junyuan and Li, Mu},
	booktitle={CVPR},
	pages={558--567},
	year={2019}
}

@inproceedings{Nichol2021ICML,
	title={Improved denoising diffusion probabilistic models},
	author={Nichol, Alexander Quinn and Dhariwal, Prafulla},
	booktitle={ICML},
	pages={8162--8171},
	year={2021}
}

@article{Zhu2025IF,
	author = {Zhu, Chunyu and Song, Xuan and Li, Yachao and Deng, Shangqi and Zhang, Tinghao},
	journal = {Information Fusion},
	pages = {103261},
	title = {{A spatial-frequency dual-domain implicit guidance method for hyperspectral and multispectral remote sensing image fusion based on Kolmogorov–Arnold Network}},
	volume = {123},
	year = {2025}
}

@article{Zhu2025TGRS,
	author = {Zhu, Chunyu and Deng, Shangqi and Song, Xuan and Li, Yachao and Wang, Qi},
	journal = {IEEE Transactions on Geoscience and Remote Sensing},
	pages = {1--15},
	title = {{Mamba Collaborative Implicit Neural Representation for Hyperspectral and Multispectral Remote Sensing Image Fusion}},
	volume = {63},
	year = {2025}
}

@inproceedings{Zheng2020CVPR-Rfmodel,
	title={What does plate glass reveal about camera calibration?},
	author={Zheng, Qian and Chen, Jinnan and Lu, Zhan and Shi, Boxin and Jiang, Xudong and Yap, Kim-Hui and Duan, Ling-Yu and Kot, Alex C},
	booktitle={CVPR},
	pages={3022--3032},
	year={2020}
}

@inproceedings{Chen2026CVPR-GFRRN,
	title={GFRRN: Explore the Gaps in Single Image Reflection Removal},
	author={Chen, Yu and He, Zewei and Liu, Xingyu and Chen, Zixuan and Lu, Zheming},
	booktitle={CVPR},
	pages={5690--5699},
	year={2026}
}

@inproceedings{Chen2024ECCV-PTTD,
	author = {Chen, Zixuan and He, Zewei and Lu, Ziqian and Sun, Xuecheng and Lu, Zhe-Ming},
	booktitle = {ECCV},
	pages = {432--449},
	title = {{Prompt-based test-time real image dehazing: a novel pipeline}},
	year = {2024}
}

@article{Chen2024TIP-DEANet,
	author = {Chen, Zixuan and He, Zewei and Lu, Zhe-Ming},
	journal = {IEEE Transactions on Image Processing},
	pages = {1002--1015},
	title = {{DEA-Net: Single Image Dehazing Based on Detail-Enhanced Convolution and Content-Guided Attention}},
	volume = {33},
	year = {2024}
}

@article{Wang2024IJCV-StableSR,
	author = {Wang, Jianyi and Yue, Zongsheng and Zhou, Shangchen and Chan, Kelvin C.K. and Loy, Chen Change},
	journal = {International Journal of Computer Vision},
	number = {12},
	pages = {5929--5949},
	title = {{Exploiting Diffusion Prior for Real-World Image Super-Resolution}},
	volume = {132},
	year = {2024}
}

@inproceedings{Zhou2022NeurIPS-CodeFormer,
	author = {Zhou, Shangchen and Chan, Kelvin C.K. and Li, Chongyi and Loy, Chen Change},
	booktitle = {NeurIPS},
	pages = {30599--30611},
	title = {{Towards Robust Blind Face Restoration with Codebook Lookup Transformer}},
	volume = {35},
	year = {2022}
}

@inproceedings{Choi2022CVPR-color,
	author = {Choi, Jooyoung and Lee, Jungbeom and Shin, Chaehun and Kim, Sungwon and Kim, Hyunwoo and Yoon, Sungroh},
	booktitle = {CVPR},
	pages = {11462--11471},
	title = {{Perception Prioritized Training of Diffusion Models}},
	year = {2022}
}

@inproceedings{Zheng2024CVPR-DiffUIR,
	title={Selective Hourglass Mapping for Universal Image Restoration Based on Diffusion Model},
	author={Zheng, Dian and Wu, Xiao-Ming and Yang, Shuzhou and Zhang, Jian and Hu, Jian-Fang and Zheng, Wei-shi},
	booktitle={CVPR},
	pages={25445--25455},
	year={2024}
}

@inproceedings{Chen2025CVPR-UniRestore,
	title={UniRestore: Unified Perceptual and Task-Oriented Image Restoration Model Using Diffusion Prior},
	author={Chen, I and Chen, Wei-Ting and Liu, Yu-Wei and Chiang, Yuan-Chun and Kuo, Sy-Yen and Yang, Ming-Hsuan and others},
	booktitle={CVPR},
	pages={17969--17979},
	year={2025}
}
\end{document}